%% file: main.tex
\documentclass[10pt,twocolumn,letterpaper]{article}

\usepackage{cvpr}
\usepackage{times}
\usepackage{epsfig}
\usepackage{graphicx}
\usepackage{amsmath}
\usepackage{amssymb}
\usepackage{helvet} 
\usepackage{courier}  
\usepackage[hyphens]{url}  
\usepackage{subfig}
\usepackage{tabularx}
\usepackage{algorithm2e}
\usepackage{algpseudocode}
\usepackage{booktabs}

\usepackage[pagebackref=true,breaklinks=true,letterpaper=true,colorlinks,bookmarks=false]{hyperref}

\cvprfinalcopy 

\def\cvprPaperID{****} 

\ifcvprfinal\fi
\begin{document}

\title{Learning When and Where to Zoom with Deep Reinforcement Learning}

\author{Burak Uzkent\\
Department of Computer Science\\
Stanford University\\
{\tt\small buzkent@cs.stanford.edu}
\and
Stefano Ermon\\
Department of Computer Science\\
Stanford University\\
{\tt\small ermon@cs.stanford.edu}
}

\newcommand{\s}[1]{{}}
\newcommand{\burak}[1]{{}}

\maketitle

\input{abstract.tex}

\input{introduction.tex}
\input{related.tex}

\input{method.tex}

\input{results.tex}

\input{conclusion.tex}

\section*{Acknowledgements}
This research was supported by Stanford’s Data for Development Initiative and NSF grants 1651565 and 1733686.

{\small
\bibliographystyle{ieee_fullname}
\bibliography{references}
}

\end{document}

%% file: abstract.tex
\begin{abstract}
While high resolution images contain semantically more useful information than their lower resolution counterparts, processing them is computationally more expensive, and in some applications, e.g. remote sensing, they can be much more expensive to acquire. For these reasons, it is desirable to develop an automatic method to selectively use high resolution data when necessary while maintaining accuracy and reducing acquisition/run-time cost. In this direction, we propose PatchDrop a reinforcement learning approach to dynamically identify when and where to use/acquire high resolution data conditioned on the paired, cheap, low resolution images. 
We conduct experiments on CIFAR10, CIFAR100, ImageNet and fMoW datasets where we use significantly less high resolution data while maintaining similar accuracy to models which use full high resolution images.
\s{rephrase this, need to explain the setting. currently doesn't make much sense}
\end{abstract}

%% file: introduction.tex
\section{Introduction}
Deep Neural Networks now achieve state-of-the-art performance in many computer vision tasks, including image recognition~\cite{deng2009imagenet}, object detection~\cite{lin2014microsoft}, and object tracking~\cite{kristan2018sixth,uzkent2014feature,uzkent2018tracking}.
However, one drawback is that they require high quality input data to perform well, and their performance drops significantly on degraded inputs, e.g., lower  resolution images~\cite{yue2016image}, lower frame rate videos~\cite{mueller2017context}, or under  distortions~\cite{sun2015learning}. 
For example, \cite{wang2016studying} studied the effect of image resolution, and  
reported a $14\%$ performance drop on CIFAR10 after downsampling images by a factor of 4. 

Nevertheless, downsampling is often performed for computational and statistical reasons~\cite{zhang2018shufflenet}.
Reducing the resolution of the inputs decreases the number of parameters, resulting in reduced computational and memory cost and mitigating overfitting~\cite{chrabaszcz2017downsampled}. Therefore, downsampling is often applied to trade off computational and memory gains with accuracy loss~\cite{ma2018shufflenet}.
However, the \emph{same} downsampling level is applied to \emph{all} the inputs.  This strategy can be suboptimal because the amount of information loss (e.g., about a label) depends on the input~\cite{gao2018dynamic}. 
Therefore, it would be desirable to build an \emph{adaptive} system to utilize a minimal amount of high resolution data while preserving accuracy.

\begin{figure*}[!h]
\centering
\subfloat[]{\includegraphics[width=0.48\textwidth]{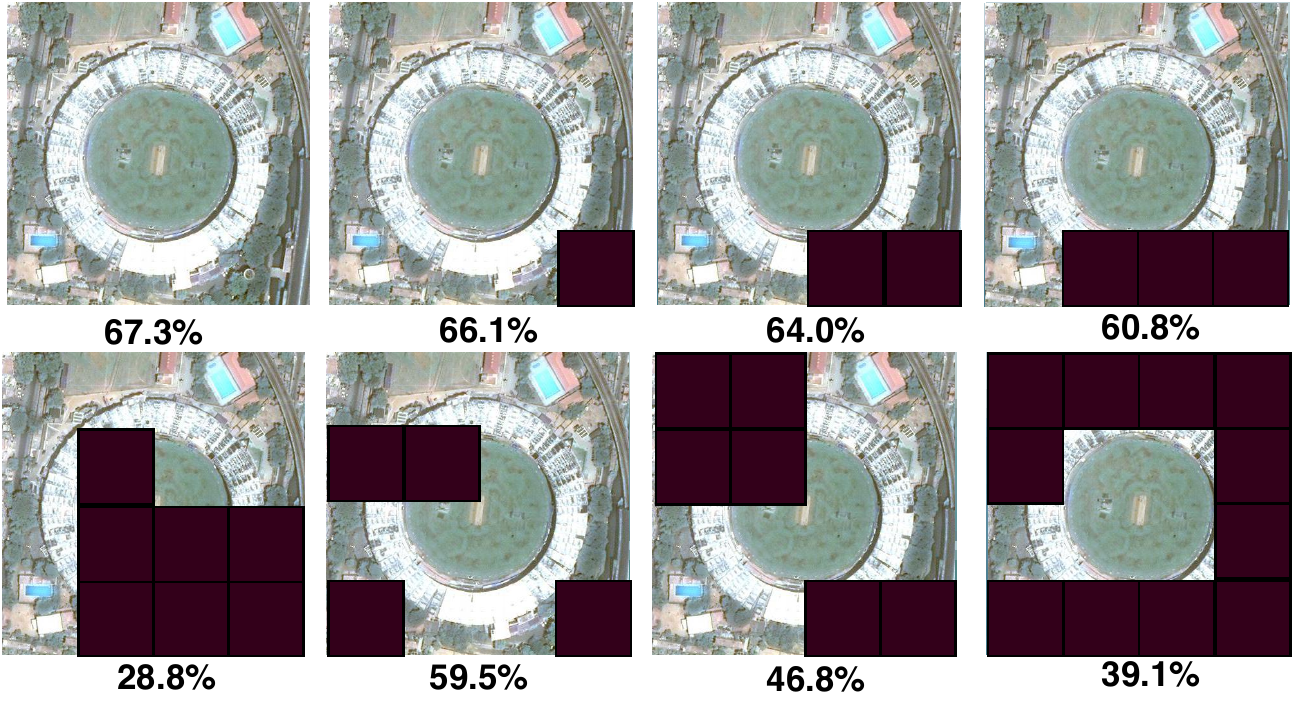}
\label{fig:patchdrop_exps}}
\subfloat[]{\includegraphics[width=0.48\textwidth]{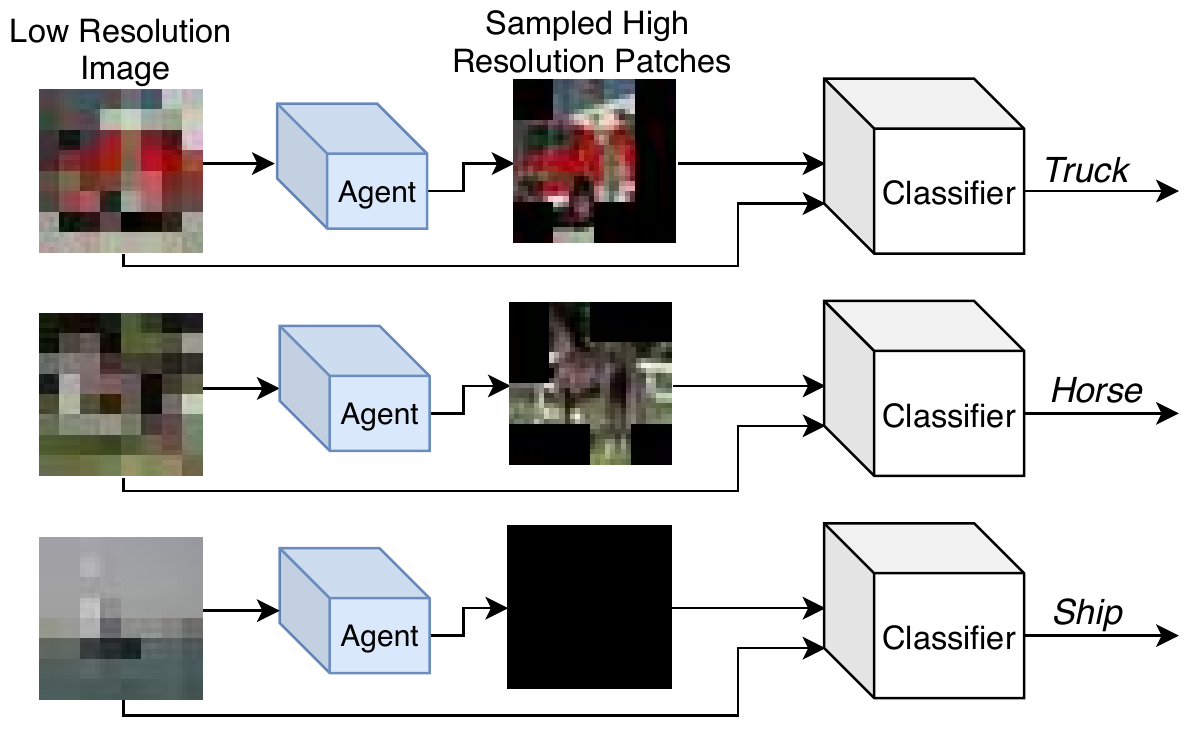}
\label{fig:patchdrop_concept}}
\caption{\textbf{Left:} shows the performance of the ResNet34 model trained on the fMoW original images and tested on images with dropped patches. The accuracy of the model goes down with the increased number of dropped patches. \textbf{Right:} shows the proposed framework which dynamically drops image patches conditioned on the low resolution images.}
\end{figure*}

In addition to computational and memory savings, an adaptive framework can also benefit application domains
where acquiring high resolution data is particularly expensive.
A prime example is remote sensing, where acquiring a high resolution (HR) satellite image is significantly more expensive than acquiring its low resolution (LR) counterpart~\cite{malarvizhi2016use,rembold2013using,ghamisi2018img2dsm}. For example, LR images with 10m-30m spatial resolution captured by Sentinel-1 satellites~\cite{geudtner2014sentinel,sarukkai2020cloud} are publicly and freely available whereas an HR image with 0.3m spatial resolution captured by DigitalGlobe satellites can cost in the order of 1,000 dollars~\cite{fisher2018impact}. This way, we can reduce the cost of deep learning models trained on satellite images for a variety of tasks, i.e., poverty prediction~\cite{sheehan2019predicting}, image recognition~\cite{uzkent2019learning,sheehan2018learning}, object tracking~\cite{uzkent2016real,uzkent2016integrating,uzkent2017aerial}. Similar examples arise in medical and scientific imaging, where acquiring higher quality images can be more expensive or even more harmful to patients~\cite{ker2017deep,kang2017deep}.

In all these settings, it would be desirable to be able to adaptively acquire only specific parts of the HR quality input. The challenge, however, is how to perform this selection automatically and efficiently, i.e., \emph{minimizing the number of acquired HR patches while retaining accuracy}.
As expected, naive strategies can be highly suboptimal. For example, randomly dropping patches of HR satellite images from the functional Map of the World (fMoW)~\cite{christie2018functional} dataset will significantly reduce accuracy of a trained network as seen in Fig.~\ref{fig:patchdrop_exps}. 
As such, an adaptive strategy must learn to identify and acquire useful patches~\cite{minut2001reinforcement} to preserve the accuracy of the network. 

To address this challenges, we propose \emph{PatchDrop}, an adaptive data sampling-acquisition scheme which only samples patches of the full HR image that are required for inferring correct decisions, as shown in Fig.~\ref{fig:patchdrop_concept}. PatchDrop uses LR versions of input images to train an agent in a reinforcement learning setting to sample HR patches only if necessary. This way, the agent learns \emph{when} and \emph{where} to zoom in the parts of the image to sample HR patches. 
PatchDrop is extremely effective on the functional Map of the World (fMoW)~\cite{christie2018functional} dataset.
Surprisingly, we show that we can use only about $40\%$ of full HR images without any significant loss of accuracy.  Considering this number, we can save in the order of 100,000 dollars when performing a computer vision task using expensive HR satellite images at global scale. We also show that PatchDrop performs well on traditional computer vision benchmarks. On ImageNet, it samples about $50\%$ of HR images on average with a minimal loss in the accuracy. On a different task, we then increase the run-time performance of patch-based CNNs, BagNets~\cite{brendel2019approximating}, by 2$\times$ by reducing the number of patches that need to be processed using  PatchDrop. Finally, leveraging the learned patch sampling policies, we generate hard positive training examples to boost the accuracy of CNNs on ImageNet and fMoW by 2-3$\%$. 

\s{mention something. then talk about compute time.
this strategy can also be used to reduce computation for bagnet...
finally, we show that this strategy can be used for data augmentation.
}

%% file: related.tex
\section{Related Work}
\textbf{Dynamic Inference with Residual Networks}
Similarly to DropOut~\cite{srivastava2014dropout},~\cite{huang2016deep} proposed a stochastic layer dropping method when training the Residual Networks~\cite{he2016deep}. The probability of survival linearly decays in the deeper layers following the hypothesis that low-level features play key roles in correct inference. Similarly, we can decay the likelihood of survival for a patch w.r.t its distance from image center based on the assumption that objects will be dominantly located in the central part of the image. Stochastic layer dropping provides only training time compression. On the other hand, \cite{wang2018skipnet,wu2018blockdrop} proposes reinforcement learning settings to drop the blocks of ResNet in both training and test time conditionally on the input image. Similarly, by replacing \emph{layers with patches}, we can drop more patches from easy samples while keeping more from ambiguous ones.

\textbf{Attention Networks}
Attention methods have been explored to localize semantically important parts of images~\cite{wang2017residual,recasens2018learning,vaswani2017attention,sun2018multi}.~\cite{wang2017residual} proposes a Residual Attention network that replaces the residual identity connections from~\cite{he2016deep} with residual attention connections. By residually learning feature guiding, they can improve recognition accuracy on different benchmarks. Similarly,~\cite{recasens2018learning} proposes a differentiable saliency-based distortion layer to spatially sample input data given a task. They use LR images in the saliency network that generates a grid highlighting semantically important parts of the image space. The grid is then applied to HR images to magnify the important parts of the image.~\cite{li2017foveanet} proposes a perspective-aware scene parsing network that locates small and distant objects. With a two branch (coarse and fovea) network, they produce coarse and fine level segmentations maps and fuse them to generate final map.~\cite{zhang2017scale} adaptively resizes the convolutional patches to improve segmentation of large and small size objects. ~\cite{lu2016adaptive} improves object detectors using pre-determined fixed anchors with adaptive ones. They divide a region into a fixed number of sub-regions recursively whenever the zoom indicator given by the network is high. 
Finally, ~\cite{pirinen2018deep} proposes a sequential region proposal network (RPN) to learn object-centric and less scattered proposal boxes for the second stage of the Faster R-CNN~\cite{ren2015faster}. These methods are tailored for certain tasks and condition the attention modules on HR images. On the other hand, we present a general framework and condition it on LR images.


\textbf{Analyzing Degraded Quality Input Signal}
There has been a relatively small volume of work on improving CNNs' performance using degraded quality input signal~\cite{li2019low}.~\cite{su2016adapting} uses knowledge distillation to train a student network using degraded input signal and the predictions of a teacher network trained on the paired higher quality signal. Another set of studies~\cite{wang2016studying,yao2019low} propose a novel method to perform domain adaptation from the HR network to a LR network.~\cite{peng2016fine} pre-trains the LR network using the HR data and finetunes it using the LR data. Other domain adaptation methods focus on person re-identification with LR images~\cite{jiao2018deep,li2015multi,wang2018resource}.
All these methods boost the accuracy of the networks on LR input data, however, they make the assumption that the quality of the input signal is fixed. 

%% file: method.tex
\section{Problem statement}
\label{sect:proposed_approach}

\begin{figure}[!h]
\centering
\vspace{0.0em}
\includegraphics[width=0.49\textwidth]{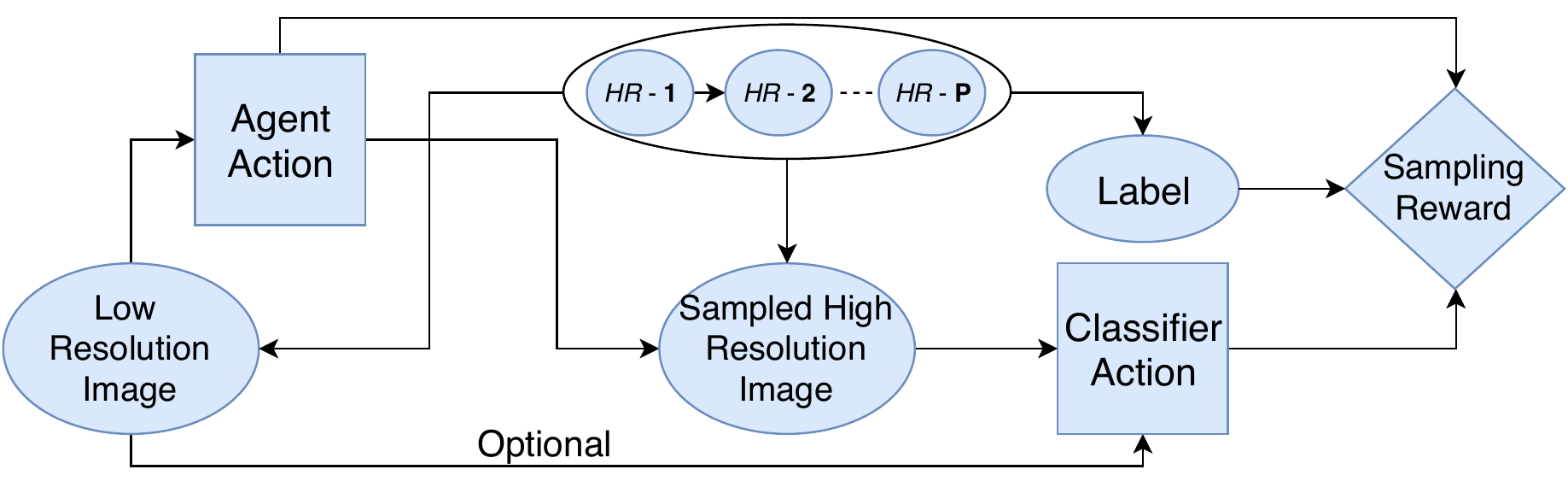}
\caption{Our Bayesian Decision influence diagram. The LR images are observed by the agent to sample HR patches. The classifier then observes the agent-sampled HR image together with the LR image to perform prediction. The ultimate goal is to choose an action to sample a masked HR image to maximize the expected utilities considering the accuracy and the cost of using/sampling HR image.}
\label{fig:task_definition}
\end{figure}

We formulate the \emph{PatchDrop} framework as a two step episodic Markov Decision Process (MDP), as shown in the influence diagram in  Fig.~\ref{fig:task_definition}. In the diagram, we represent the random variables with a circle, actions with a square, and utilities with a diamond. A high spatial resolution image, $x_{h}$, is formed by equal size patches with zero overlap $x_{h} = (x_{h}^{1}, x_{h}^{2},\cdots, x_{h}^{P})$, where $P$ represents the number of patches. In contrast with traditional computer vision settings, $x_{h}$ is latent, \textit{i.e.}, it is  \emph{not observed by the agent}.
$y \in \{1, \cdots, N\}$ is a categorical random variable representing the (unobserved) label associated with $x_{h}$, where $N$ is the number of classes. The random variable \emph{low spatial resolution image}, $x_{l}$, is the lower resolution version of $x_{h}$. $x_{l}$ is initially observed by the agent in order to choose the binary action array, $\mathbf{a_{1}} \in \{0,1\}^{P}$, where $\mathbf{a_1^p} =1$ means that the agent would like to sample the $p$-th HR patch $x_{h}^{p}$.  
We define the patch sampling policy model parameterized by $\theta_{p}$, as
\begin{equation}
    \pi_{1}(\mathbf{a_1} | x_{l};\theta_{p}) = p(\mathbf{a_{1}}|x_{l};\theta_{p}),
\end{equation}
where $\pi_{1}(x_{l};\theta_{p})$ is a function mapping the observed LR image to a probability distribution over the patch sampling action $\mathbf{a_{1}}$. 
Next, the random variable \emph{masked HR image}, $x_{h}^{m}$, is formed using $\mathbf{a_{1}^{p}}$ and $x_{h}^{p}$, with the masking operation formulated as $x_{h}^{m} = x_{h}\bigodot \mathbf{a_1}$.
The first step of the MDP can be modeled with a joint probability distribution over the random variables, $x_{h}$, $y$, $x_{h}^{m}$, and $x_{l}$, and action $\mathbf{a_{1}}$, as
\begin{align}
    p(x_{h}, x_{h}^{m}, x_{l}, y, \mathbf{a_1}) =  p(x_{h})p(y|x_{h})p(x_{l}|x_{h}) \nonumber \\ \cdot p(\mathbf{a_{1}}|x_{l};\theta_{p})  p(x_{h}^{m}|\mathbf{a_{1}}, x_{h}).
\end{align}


In the second step of the MDP, the agent observes the random variables, $x_{h}^{m}$ and $x_{l}$, and chooses an action $\mathbf{a_{2}} \in \{1, \cdots, N\}$. We then define the class prediction policy as follows:
\begin{equation}
    \pi_{2}(\mathbf{a_2} | x_{h}^{m}, x_{l};\theta_{cl}) = p(\mathbf{a_{2}}|x_{h}^{m}, x_{l};\theta_{cl}),
\end{equation}
where $\pi_{2}$ represents a classifier network parameterized by $\theta_{cl}$. 
The overall objective, $J$, is then defined as maximizing the expected utility, $R$
represented by
\begin{equation}
\max_{\theta_{p}, \theta_{cl}} J(\theta_{p}, \theta_{cl}) = \mathbb{E}_p[R(\mathbf{a_{1}}, \mathbf{a_{2}}, y)],
\label{Eq:Cost_Function}
\end{equation}
where the utility
depends on $\mathbf{a_{1}}$, $\mathbf{a_{2}}$, and $y$. The reward penalizes the agent for selecting a large number of high-resolution patches (e.g., based on the norm of $\mathbf{a}_{1}$) and includes a classification loss evaluating the accuracy of $\mathbf{a}_{2}$ given the true label $y$ (e.g., cross-entropy or 0-1 loss).

\section{Proposed Solution}
\subsection{Modeling the Policy Network and Classifier}
\begin{figure*}[!h]
\centering
\includegraphics[width=0.98\textwidth]{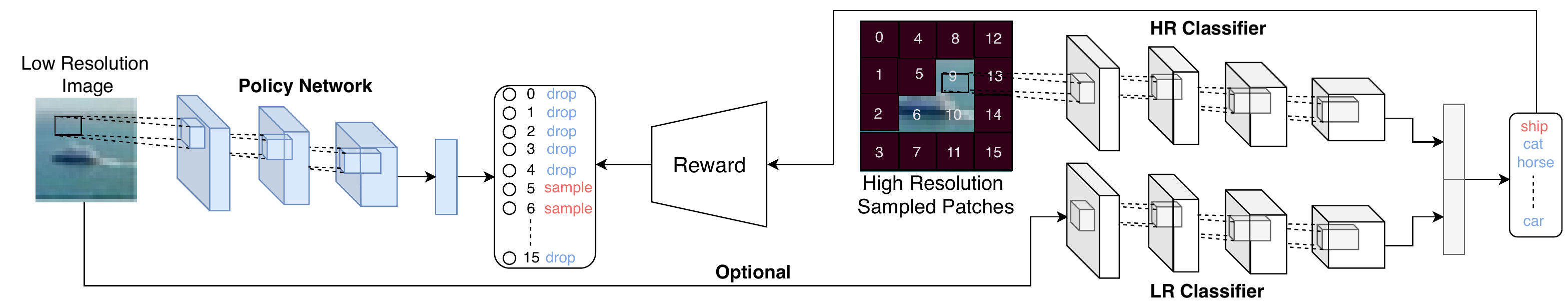}
\caption{The workflow of the \emph{PatchDrop} formulated as a two step episodic MDP. The agent chooses actions conditioned on the LR image, and only agent sampled HR patches together with LR images are jointly used by the two-stream classifier network. We note that the LR network can be disconnected from the pipeline to only rely on selected HR patches to perform classification. When disconnecting LR network, the policy network samples more patches to maintain accuracy.}
\label{fig:patchdrop_workflow}
\end{figure*}

In the previous section, we formulated the task of \emph{PatchDrop} as a two step episodic MDP similarly to~\cite{uzkent2020efficient}.
Here, we detail the action space and how the policy distributions for $\mathbf{a_{1}}$ and $\mathbf{a_{2}}$ are modelled. To represent our discrete action space for $\mathbf{a_{1}}$, we divide the image space into equal size patches with no overlaps, resulting in $P$ patches, as shown in Fig.~\ref{fig:patchdrop_workflow}. In this study, we use $P=16$ regardless of the size of the input image and leave the task of choosing variable size bounding boxes as a future work. In the first step of the two step MDP, the policy network, $f_{p}$, \emph{outputs the probabilities for all the actions at once} after observing $x_{l}$. An alternative approach could be in the form of a Bayesian framework where $\mathbf{a_{1}^{p}}$ is conditioned on $\mathbf{a_{1}^{1:p-1}}$~\cite{gao2018dynamic,pirinen2018deep}. However, the proposed concept of \emph{outputting all the actions at once} provides a more efficient decision making process for patch sampling.

In this study, we model the action likelihood function of the policy network, $f_{p}$, by multiplying the probabilities of the individual high-resolution patch selections, represented by patch-specific Bernoulli distributions as follows:
\begin{equation}
  \pi_{1}({\mathbf{a}_{1}|x_{l},\theta_{p}) = \prod_{p=1}^{P} s_{p}^{\mathbf{a_{1}^{p}}}(1- s_{p})^{(1-\mathbf{a_{1}^{p}})}},
\label{eq:action_likelihood}
\end{equation}
where $s_{p}$ represents the prediction vector formulated as
\begin{equation}
    s_{p} = f_{p}(x_{l};\theta_{p}).
\label{eq:policy_network}
\end{equation}
To get probabilistic values, $s_{p} \in [0,1]$, we use a sigmoid function on the final layer of the policy network.

The next set of actions, $\mathbf{a_{2}}$, is chosen by the classifier, $f_{cl}$, using the sampled HR image $x_{h}^{m}$ and the LR input $x_{l}$. The upper stream of the classifier, $f_{cl}$, uses the sampled HR images, $x_{h}^{m}$, whereas the bottom stream uses the LR images, $x_{l}$, as shown in Fig.~\ref{fig:patchdrop_workflow}.
Each one outputs probability distributions, $s_{cl}^{l}$ and $s_{cl}^{h^{m}}$, for class labels using a softmax layer. We then compute the weighted sum of predictions via
\begin{equation}
    s_{cl} = (S/P)s_{cl}^{h^{m}} + (1-S/P)s_{cl}^{l},
\label{eq:weighted_sum}
\end{equation}
where $S$ represents the number of sampled patches. To form $\mathbf{a_{2}}$, we use the maximally probable class label: \textit{i.e.}, $\mathbf{a_{2}^{j}}=1$ if $s_{cl}^{j}=\max(s_{cl})$ and $\mathbf{a_{2}^{j}}=0$ otherwise where $j$ represents the class index. In this set up, if the policy network samples \emph{no HR patch}, we completely rely on the LR classifier, and the impact of the HR classifier increases linearly with the number of sampled patches.

\subsection{Training the PatchDrop Network} 
After defining the two step MDP and modeling the policy and classifier networks, we detail the training procedure of \emph{PatchDrop}. The goal of training is to learn the optimal parameters of $\theta_{p}$ and $\theta_{cl}$. 
Because the actions are discrete, we cannot use the reparameterization trick to optimize the objective w.r.t. $\theta_{p}$.
To optimize the parameters $\theta_{p}$ of $f_{p}$, we need to use model-free reinforcement learning algorithms such as 
Q-learning~\cite{Watkins92q-learning} and 
policy gradient~\cite{sutton2018reinforcement}. Policy gradient is more suitable in our scenario since the number of unique actions the policy network can choose is $2^{P}$ and increases exponentially with $P$. For this reason, we use the $REINFORCE$ method~\cite{sutton2018reinforcement} to optimize the objective w.r.t $\theta_{p}$ using
\begin{equation}
\nabla_{\theta_{p}}J = \mathbb{E}[R(\mathbf{a_{1}}, \mathbf{a_{2}}, y)\nabla_{\theta_{p}} \log \pi_{\theta_{p}}(\mathbf{a_{1}}|x_{l})].
\label{eq:policy_gradient}
\end{equation}
Averaging across a mini-batch via Monte-Carlo sampling produces an unbiased estimate of the expected value, but with potentially large variance. Since this can lead to an unstable training process~\cite{sutton2018reinforcement}, we replace $R(\mathbf{a_{1}}, \mathbf{a_{2}}, y)$ in Eq.~\ref{eq:policy_gradient} with the advantage function to reduce the variance:
\begin{equation}
\nabla_{\theta_{p}}J = \mathbb{E}[A \sum_{p=1}^{P} \nabla_{\theta_{p}}\log(s_{p}\mathbf{a_{1}^{p}}+(1-s_{p})(1-\mathbf{a_{1}^{p}}))],
\end{equation}
\begin{equation}
    A(\mathbf{a_{1}}, \mathbf{\hat{a}_{1}}, \mathbf{a_{2}}, \mathbf{\hat{a}_{2}}) = R(\mathbf{a_{1}}, \mathbf{a_{2}}, y) - R(\mathbf{\hat{a}_{1}}, \mathbf{\hat{a}_{2}}, y),
\end{equation}
where $\mathbf{\hat{a}_{1}}$ and $\mathbf{\hat{a}_{2}}$ represent the baseline action vectors. To get $\mathbf{\hat{a}_{1}}$, we use the most likely action vector proposed by the policy network: \textit{i.e.}, $\mathbf{a_{1}^{i}}=1$ if $s_{p}^{i}>0.5$ and $s_{p}^{i}=0$ otherwise. The classifier, $f_{cl}$, then observes $x_{l}$ and $\hat{x}_{h}^{m}$ sampled using $\mathbf{\hat{a}_{1}}$, on two branches and outputs the predictions, $\hat{s}_{cl}$, from which we get $\mathbf{\hat{a}_{2}^{j}}$: \textit{i.e.}, $\mathbf{\hat{a}_{2}^{j}}=1$ if $\hat{s}_{cl}^{j}=\max(\hat{s}_{cl})$ and $\mathbf{\hat{a}_{2}^{j}}=0$ otherwise where $j$ represent the class index. The advantage function assigns the policy network \emph{a positive value} only when the action vector sampled from Eq.~\ref{eq:action_likelihood} produces higher reward than the action vector with maximum likelihood, which is known as a self-critical baseline~\cite{rennie2017self}. 

Finally, in this study we use the temperature scaling method~\cite{sutton2018reinforcement} to encourage exploration during training time by bounding the probabilities of the policy network as
\begin{equation}
s_{p} = \alpha s_{p} + (1-\alpha)(1-s_{p}),
\end{equation}
where $\alpha \in [0, 1]$.

\textbf{Pre-training the Classifier}
After formulating our reinforcement learning setting for training the policy network, we first pre-train the two branches of $f_{cl}$, $f_{cl}^{h}$ and $f_{cl}^{l}$, on $x_{h} \in \mathcal{X}_{h}$ and $x_{l} \in \mathcal{X}_{l}$. We assume that $\mathcal{X}_{h}$ is observable in the training time. The network trained on $\mathcal{X}_{h}$ can perform reasonably (Fig.~\ref{fig:patchdrop_exps}) when the patches are dropped at test time with a \emph{fixed policy}, forming $x_{h}^{m}$. We then use this observation to pre-train the policy network, $f_{p}$, to dynamically learn to drop patches while keeping the parameters of $f_{cl}^{h}$ and $f_{cl}^{l}$ fixed.

\textbf{Pre-training the Policy Network (Pt)}
After training the two streams of the classifier, $f_{cl}$, we pre-train the policy network, $f_{p}$, using the proposed reinforcement learning setting while fixing the parameters of $f_{cl}$. In this step, we only use $f_{cl}^{h}$ to estimate the expected reward when learning $\theta_{p}$. This is because we want to train the policy network to understand which patches contribute most to correct decisions made by the HR image classifier, as shown in Fig.~\ref{fig:patchdrop_exps}. 

\begin{algorithm}[!h]
\hrule
\SetAlgoLined
\KwIn{Input($\mathcal{X}_{l}$, $\mathcal{Y}$, $\mathcal{C}$)\quad $\mathcal{X}_{l}=\{x_{l}^{1}, x_{l}^{2}, ..., x_{l}^{N}\}$}
\For{$k\gets 0\:to\:K_{1}$}{
    $s_{p} \gets f_{p}(x_{l};\theta_{p})$ \\
    $s_{p} \gets \alpha + (1-s_{p})(1-\alpha)$ \\
    $\mathbf{a_{1}} \sim \pi_{1}(a_{1}|s_p)$ \\
    \BlankLine
    $x_{h}^{m} = x_{h}\bigodot\:\mathbf{a_{1}}$ \\
    \BlankLine
    $\mathbf{a_2} \gets f_{cl}^{h}(x_{h}^{m};\theta_{cl}^{h})$ \\
    \BlankLine
    \textbf{Evaluate Reward} $R(\mathbf{a_{1}}, \mathbf{a_{2}}, y)$ \\
    \BlankLine
    $\theta_{p} \gets \theta_{p} + \nabla{\theta_{p}}$
}
\For{$k \gets 0\:to\:K_{2}$}{
    \textbf{Jointly Finetune} $\theta_{cl}^{h}$ and $\theta_{p}$ using $f_{cl}^{h}$
}
\For{$k\gets 0\:to\:K_{3}$}{
    \textbf{Jointly Finetune $\theta_{cl}^{h}$} and $\theta_{p}$ using $f_{cl}^{h}$ and $f_{cl}^{l}$
}
\hrule
\caption{PatchDrop Pseudocode}
\label{alg:PatchDrop_Algorithm}
\end{algorithm}

\textbf{Finetuning the Agent and HR Classifier (Ft-1)}
To further boost the accuracy of the policy network, $f_{p}$, we jointly finetune the policy network and HR classifier, $f_{cl}^{h}$. This way, the HR classifier can adapt to the sampled images, $x_{h}^{m}$, while the policy network learns new policies in line with it. The LR classifier, $f_{cl}^{l}$, is not included in this step. 

\textbf{Finetuning the Agent and HR Classifier (Ft-2)}
In the final step of the training stage, we jointly finetune the policy network, $f_{p}$, and $f_{cl}^{h}$ \emph{with the addition} of $f_{cl}^{l}$ into the classifier $f_{cl}$. This way, the policy network can learn policies to drop further patches with the existence of the LR classifier. We combine the HR and LR classifiers using Eq.~\ref{eq:weighted_sum}. Since the input to $f_{cl}^{l}$ does not change, we keep $\theta_{cl}^{l}$ fixed and only update $\theta_{cl}^{h}$ and $\theta_{p}$. The algorithm for the \emph{PatchDrop} training stage is shown in Alg.~\ref{alg:PatchDrop_Algorithm}. Upon publication, we will release the code to train and test PatchDrop.

%% file: results.tex
\section{Experiments}
\subsection{Experimental Setup}
\textbf{Datasets and Metrics} We evaluate \emph{PatchDrop} on the following datasets: (1) CIFAR10, (2) CIFAR100, (3) ImageNet~\cite{deng2009imagenet} and (4) functional map of the world (fMoW)~\cite{christie2018functional}. To measure its performance, we use image recognition accuracy and the number of dropped patches (cost).

\textbf{Implementation Details}
In CIFAR10/CIFAR100 experiments, we use a ResNet8 for the policy and ResNet32 for the classifier networks. The policy and classifier networks use 8$\times$8px and 32$\times$32px images. In ImageNet/fMoW, we use a ResNet10 for the policy network and ResNet50 for the classifier. The policy network uses 56$\times$56px images whereas the classifier uses 224$\times$224px images. We initialize the weights of the LR classifier with HR classifier~\cite{peng2016fine} and use Adam optimizer in all our experiments~\cite{kingma2014adam}. Finally, initially we set the exploration/exploitation parameter, $\alpha$, to 0.7 and increase it to 0.95 linearly over time. Our PatchDrop implementation can be found in our GitHub repository~\footnote{https://github.com/ermongroup/PatchDrop}.

\textbf{Reward Function} We choose $R=1 - \left(\dfrac{|\mathbf{a_{1}}|_{1}}{P}\right)^{2}$ if \text{$y=\hat{y}(\mathbf{a_{2}})$} and -$\sigma$ otherwise as a reward. Here, $\hat{y}$ and $y$ represent the predicted class by the classifier after the observation of $x_{h}^{m}$ and $x_{l}$ and the true class, respectively.
The proposed reward function quadratically increases the reward w.r.t the number of dropped patches. To adjust the trade-off between accuracy and the number of sampled patches, we introduce $\sigma$ and setting it to a large value encourages the agent to sample more patches to preserve accuracy.

\subsection{Baseline and State-of-The-Art Models}
\textbf{No Patch Sampling/No Patch Dropping} In this case, we simply train a CNN on LR or HR images with cross-entropy loss without any domain adaptation and test it on LR or HR images. We call them LR-CNN and HR-CNN.

\textbf{Fixed and Stochastic Patch Dropping} 
We propose two baselines that sample central patches along the horizontal and vertical axes of the image space and call them Fixed-H and Fixed-V. 
We list the sampling priorities for the patches in this order 5,6,9,10,13,14,1,2,0,3,4,7,8,11,15 for \emph{Fixed-H}, and 4,5,6,7,8,9,10,11,12,13,14,15,0,1,2,3 for \emph{Fixed-V}. The patch IDs are shown in Fig.~\ref{fig:patchdrop_workflow}. 
Using a similar hypothesis, we then design a stochastic method that decays the survival likelihood of a patch w.r.t the euclidean distance from the center of the patch $p$ to the image center.
\begin{figure*}[!h]
\centering
\includegraphics[width=0.98\textwidth]{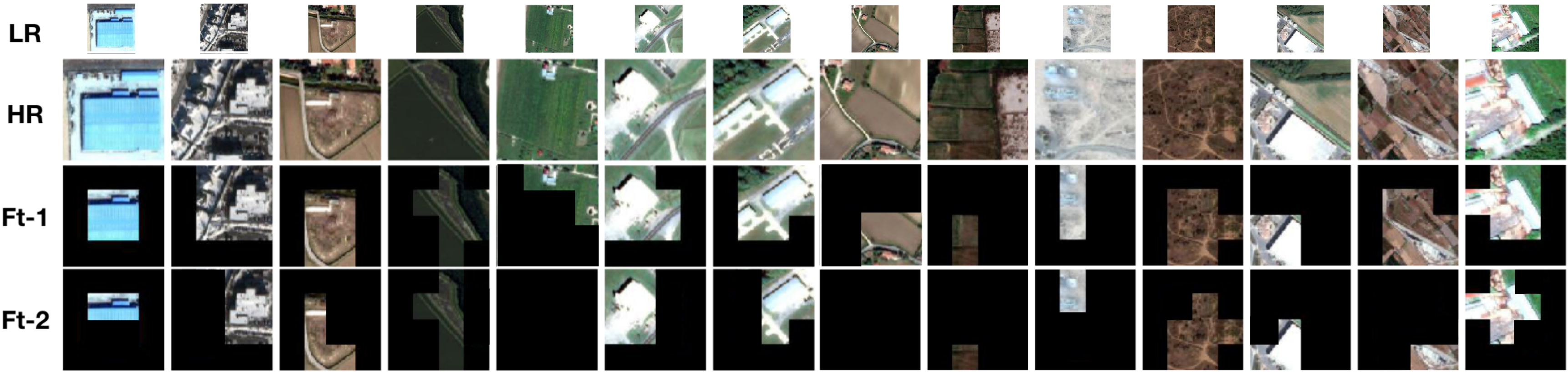}
\caption{Policies learned on the fMoW dataset. In columns 5 and 8, Ft-2 model does not sample any HR patches and the LR classifier is used. Ft-1 model samples more patches as it does not utilize LR classifier.}
\label{fig:fmow_visuals}
\end{figure*}

\textbf{Super-resolution} We use SRGAN~\cite{ledig2017photo} to learn to upsample LR images and use the SR images in the downstream tasks. This method only improves accuracy and increases computational complexity since SR images have the same number of pixels with HR images.

\textbf{Attention-based Patch Dropping} 
In terms of the state-of-the art models, we first compare our method to the Spatial Transformer Network (STN) by~\cite{recasens2018learning}. We treat their saliency network as the policy network and sample the top $S$ activated patches to form masked images for classifier.

\textbf{Domain Adaptation} Finally, we use two of the state-of-the-art domain adaptation methods by~\cite{wang2016studying,su2016adapting} to improve recognition accuracy on LR images. These methods are based on Partially Coupled Networks (PCN), and Knowledge Distillation (KD)~\cite{hinton2015distilling}.

The LR-CNN, HR-CNN, PCN, KD, and SRGAN are standalone models and always use full LR or HR image. For this reason, we have same values for them in Pt, Ft-1, and Ft-2 steps and show them in the upper part of the tables.
\subsection{Experiments on fMoW}
One application domain of the PatchDrop is remote sensing where LR images are significantly cheaper than HR images. In this direction, we test the PatchDrop on functional Map of the World~\cite{christie2018functional} consisting of HR satellite images. We use 350,000, 50,000 and 50,000 images as training, validation and test images. After training the classifiers, we pre-train the policy network for 63 epochs with a learning rate of 1e-4 and batch size of 1024. Next, we finetune (Ft-1 and Ft-2) the policy network and HR classifiers with the learning rate of 1e-4 and batch size of 128. Finally, we set $\sigma$ to 0.5, 20, and 20 in the pre-training, and fine-tuning steps.
\begin{table}[!b]
\centering
\resizebox{0.99\columnwidth}{!}{%
\begin{tabular}{@{}lllllll@{}}
\toprule
 & \multicolumn{1}{c}{\begin{tabular}[c]{@{}c@{}}Acc. (\%)\\ (Pt)\end{tabular}} & S & \multicolumn{1}{c}{\begin{tabular}[c]{@{}c@{}}Acc. (\%)\\ (Ft-1)\end{tabular}} & S & \multicolumn{1}{c}{\begin{tabular}[c]{@{}c@{}}Acc. (\%)\\ (Ft-2)\end{tabular}} & S\\ \midrule
LR-CNN & 61.4 & 0 & 61.4 & 0 & 61.4 & 0 \\ 
SRGAN~\cite{ledig2017photo} & 62.3 & 0 & 62.3 & 0 & 62.3 & 0\\
KD~\cite{su2016adapting} & 63.1 & 0 & 63.1 & 0 & 63.1 & 0\\
PCN~\cite{wang2016studying} & 63.5 & 0 & 63.5 & 0 & 63.5 & 0\\
HR-CNN & 67.3 & 16 & 67.3 & 16 & 67.3 & 16  \\ \bottomrule
Fixed-H & 47.7 & 7 & 63.3 & 6 & 64.9 & 6\\
Fixed-V & 48.3 & 7 & 63.2 & 6 & 64.7 & 6\\
Stochastic & 29.1 & 7 & 57.1 & 6 & 63.6 & 6\\
STN~\cite{recasens2018learning} & 46.5 & 7 & 61.8 & 6 & 64.8 & 6\\
\textbf{PatchDrop} & \textbf{53.4} & \textbf{7} & \textbf{67.1} & \textbf{5.9} & \textbf{68.3} & \textbf{5.2} \\ 
\bottomrule
\end{tabular}}
\vspace{0.3em}
\caption{The performance of the proposed \emph{PatchDrop} and baseline models on the fMoW dataset. $S$ represents the average number of sampled patches. Ft-1 and Ft-2 represent the finetuning steps with single and two stream classifiers.}
\label{table:fmow}
\end{table}
\begin{figure}[!b]
\centering
\includegraphics[width=0.214\textwidth]{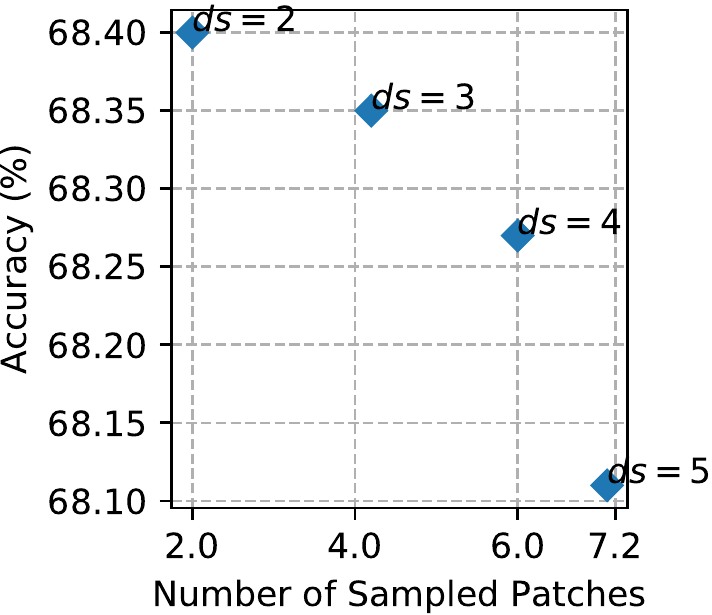}
\includegraphics[width=0.21\textwidth]{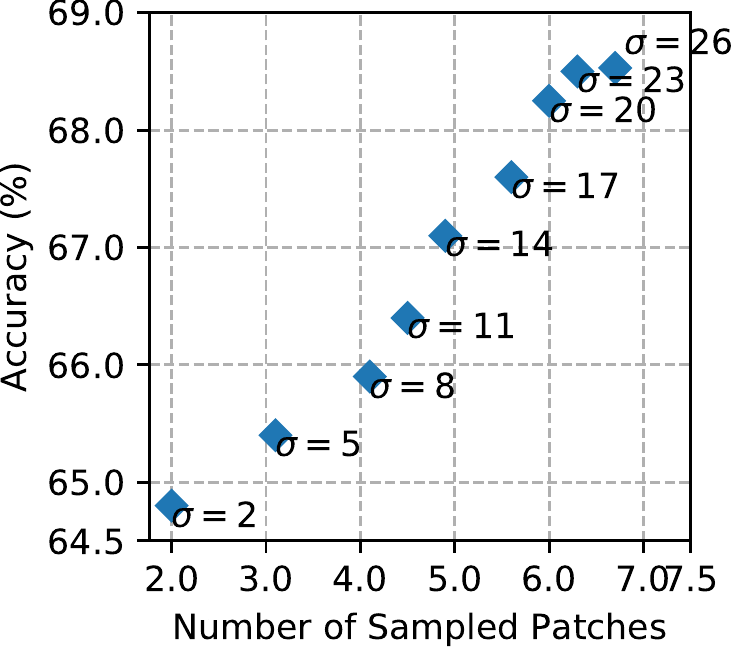}
\caption{\textbf{Left:} The accuracy and number of sampled patches by the policy network w.r.t downsampling ratio used to get LR images for the policy network and classifier. \textbf{Right:} The accuracy and number of sampled patches w.r.t to $\sigma$ parameter in the reward function in the joint finetuning steps ($ds$=4). It is set to 0.5 in the pre-training step.
}
\label{fig:analysis_pn_fmow}
\end{figure}

As seen in Table~\ref{table:fmow}, PatchDrop samples only about $40\%$ of each HR image on average while increasing the accuracy of the network using the full HR images to $68.3\%$. Fig.~\ref{fig:fmow_visuals} shows some examples of how the policy network chooses actions conditioned on the LR images. When the image contains a field with uniform texture, the agent samples a small number of patches, as seen in columns 5, 8, 9 and 10. On the other hand, it samples patches from the buildings when the ground truth class represents a building, as seen in columns 1, 6, 12, and 13.

\begin{figure*}[!h]
\centering
\includegraphics[width=0.94\textwidth]{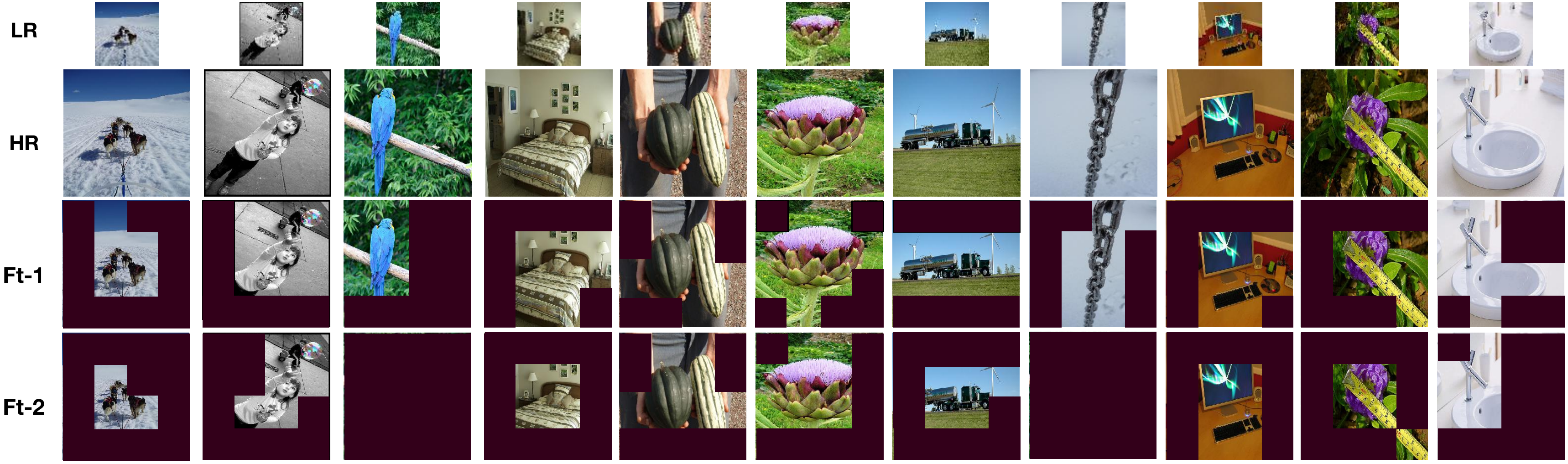}
\caption{Policies learned on ImageNet. In columns 3 and 8, Ft-2 model does not sample any HR patches and the LR classifier is used. Ft-1 model samples more patches as it does not use the LR classifier.} 
\label{fig:imgnet_visuals}
\end{figure*}
Also, we perform experiments with different downsampling ratios and $\sigma$ values in the reward function. This way, we can observe the trade-off between the number of sampled patches and accuracy. As seen in Fig.~\ref{fig:analysis_pn_fmow}, as we increase the downsampling ratio we zoom into more patches to maintain accuracy. On the other hand, with increasing $\sigma$, we zoom into more patches as larger $\sigma$ value penalizes the policies resulting in unsuccessful classification.

\begin{table*}[!h]
\centering
\resizebox{0.98\textwidth}{!}{%
\begin{tabular}{@{}lllllllllllll@{}}
\toprule
 & \multicolumn{4}{c}{CIFAR10} & \multicolumn{4}{c}{CIFAR100} &  \multicolumn{4}{c}{ImageNet} \\ \midrule
 & \begin{tabular}[c]{@{}l@{}}Acc. (\%)\\ (Pt)\end{tabular} & \begin{tabular}[c]{@{}l@{}}Acc. (\%)\\ (Ft-1)\end{tabular} & \begin{tabular}[c]{@{}l@{}}Acc. (\%)\\ (Ft-2)\end{tabular} & \begin{tabular}[c]{@{}l@{}}S\\ (Pt,Ft-1,Ft-2)\end{tabular} & \begin{tabular}[c]{@{}l@{}}Acc. (\%)\\ (Pt)\end{tabular} & \begin{tabular}[c]{@{}l@{}}Acc. (\%)\\ (Ft-1)\end{tabular} & \begin{tabular}[c]{@{}l@{}}Acc. (\%)\\ (Ft-2)\end{tabular} & \begin{tabular}[c]{@{}l@{}}S\\ (Pt,Ft-1,Ft-2)\end{tabular} & \begin{tabular}[c]{@{}l@{}}Acc. (\%)\\ (Pt)\end{tabular} & \begin{tabular}[c]{@{}l@{}}Acc. (\%)\\ (Ft-1)\end{tabular} & \begin{tabular}[c]{@{}l@{}}Acc. (\%)\\ (Ft-2)\end{tabular} & \begin{tabular}[c]{@{}l@{}}S\\ (Pt,Ft-1,Ft-2)\end{tabular} \\
LR-CNN & 75.8 & 75.8 & 75.8 & 0,0,0 & 55.1 & 55.1 & 55.1 & 0,0,0 & 58.1 & 58.1 & 58.1 & 0,0,0\\ 
SRGAN~\cite{ledig2017photo} & 78.8 & 78.8 & 78.8 & 0,0,0 & 56.1 & 56.1 & 56.1 & 0,0,0 & 63.1 & 63.1 & 63.1 & 0,0,0 \\
KD~\cite{su2016adapting} & 81.8 & 81.8 & 81.8 & 0,0,0 & 61.1 & 61.1 & 61.1 & 0,0,0 & 62.4 & 62.4 & 62.4 & 0,0,0  \\
PCN~\cite{su2016adapting} & 83.3 & 83.3 & 83.3 & 0,0,0 & 62.6 & 62.6 & 62.6 & 0,0,0 & 63.9 & 63.9 & 63.9 & 0,0,0  \\
HR-CNN & 92.3 & 92.3 & 92.3 & 16,16,16 & 69.3 & 69.3 & 69.3 & 16,16,16 & 76.5 & 76.5 & 76.5 & 16,16,16\\
 \bottomrule
Fixed-H & 71.2 & 83.8 & 85.2 & 9,8,7 & 48.5 & 65.8 & 67.0 & 9,10,10 & 48.8 & 68.6 & 70.4 & 10,9,8\\
Fixed-V & 64.7 & 83.4 & 85.1 & 9,8,7 & 46.2 & 65.5 & 67.2 & 9,10,10 & 48.4 & 68.4 & 70.8 & 10,9,8 \\
Stochastic & 40.6 & 82.1 & 83.7 & 9,8,7 & 27.6 & 63.2 & 64.8 & 9,10,10 & 38.6 & 66.2 & 68.4 & 10,9,8 \\
STN~\cite{recasens2018learning} & 66.9 & 85.2 & 87.1 & 9,8,7 & 41.1 & 64.3 & 66.4 & 9,10,10 & 58.6 & 69.4 & 71.4 & 10,9,8  \\
\textbf{PatchDrop} & \textbf{80.6} & \textbf{91.9} & \textbf{91.5} & \textbf{8.5,7.9,6.9} & \textbf{57.3} & \textbf{69.3} & \textbf{70.4} & \textbf{9,9.9,9.1} & \textbf{60.2} & \textbf{74.9} & \textbf{76.0} & \textbf{10.1},\textbf{9.1},\textbf{7.9}\\ 
\bottomrule
\end{tabular}}
\vspace{0.3em}
\caption{The results on CIFAR10, CIFAR100 and ImageNet datasets. $S$ represents the average number of sampled patches per image. The Pt, Ft-1 and Ft-2 represent the pre-training and finetuning steps with single and two stream classifiers.}
\label{table:CIFAR10}
\end{table*}
\textbf{Experiments on CIFAR10/CIFAR100} Although CIFAR datasets already consists of LR images, we believe that conducting experiments on standard benchmarks is useful to characterize the model. For CIFAR10, after training the classifiers, we pre-train the policy network with a batch size of 1024 and learning rate of 1e-4 for 3400 epochs. In the joint finetuning stages, we keep the learning rate, reduce the batch size to 256, and train the policy and HR classifier networks for 1680 and 990 epochs, respectively. $\sigma$ is set to -0.5 in the pre-training stage and -5 in the joint finetuning stages whereas $\alpha$ is tuned to 0.8. Our CIFAR100 methods are similar to the CIFAR10 ones, including hyper-parameters.


As seen in Table~\ref{table:CIFAR10}, \emph{PatchDrop} drops about $56\%$ of the patches in the original image space in CIFAR10, all the while with minimal loss in the overall accuracy. In the case of CIFAR100, we observe that it samples 2.2 patches more than the CIFAR10 experiment, on average, which might be due to higher complexity of the CIFAR100 dataset. 

\textbf{Experiments on ImageNet}
Next, we test the PatchDrop on  ImageNet Large Scale Visual Recognition
Challenge 2012 (ILSVRC2012)~\cite{russakovsky2015ilsvrc}.
It contains 1.2 million, 50,000, and 150,000 training, validation and test images. For augmentation, we use randomly cropping 224$\times$224px area from the 256$\times$256px images and perform horizontal flip augmentation.
After training the classifiers, we pre-train the policy network for 95 epochs with a learning rate of 1e-4 and batch size of 1024. We then perform the first fine-tuning stage and jointly finetune the HR classifier and policy network for 51 epochs with the learning rate of 1e-4 and batch size of 128. Finally, we add the LR classifier and jointly finetune the policy network and HR classifier for 10 epochs with the same learning rate and batch size. We set $\sigma$ to 0.1, 10, and 10 for pre-training and fine-tuning steps.

As seen in Table~\ref{table:CIFAR10}, we can maintain the accuracy of the HR classifier while dropping $56\%$ and $50\%$ of the patches with the Ft-1 and Ft-2 model. Also, we show the learned policies on ImageNet in Fig.~\ref{fig:imgnet_visuals}. The policy network decides to sample no patch when the input is relatively easier as in column 3, and 8. 

\begin{figure}[!t]
\centering
\includegraphics[width=0.22\textwidth]{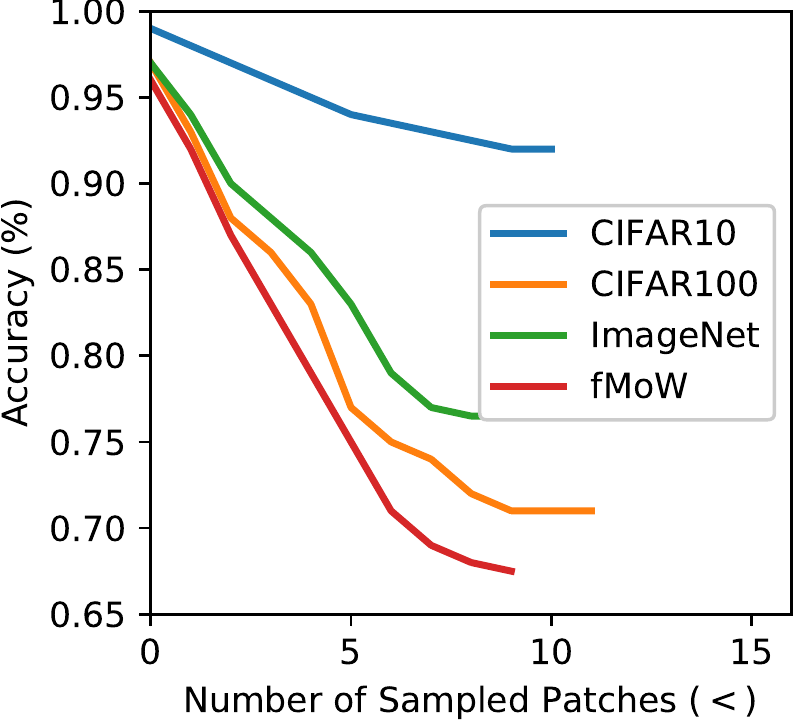}
\includegraphics[width=0.22\textwidth]{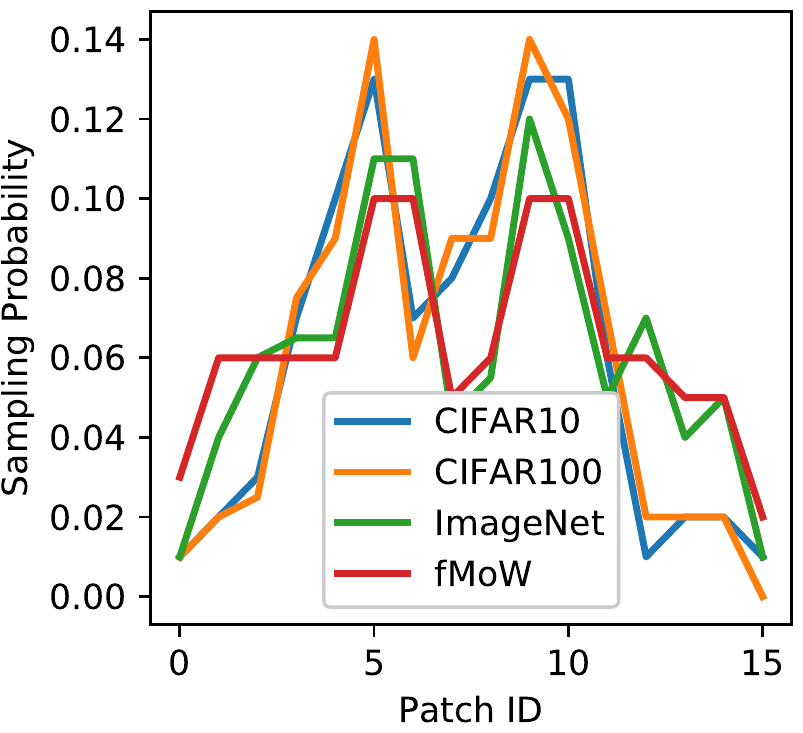}
\caption{\textbf{Left:} The accuracy w.r.t the average number of sampled patches by the policy network. 
\textbf{Right:} Sampling probability of the patch IDs (See Fig.~\ref{fig:patchdrop_workflow} for IDs). 
}
\label{fig:analysis_pn}
\end{figure}

\textbf{Analyzing Policy Network's Actions}
To better understand the sampling actions of policy network, we visualize the accuracy of the classifier w.r.t the number of sampled patches as shown in Fig.~\ref{fig:analysis_pn} (left). Interestingly, the accuracy of the classifier is \emph{inversely} proportional to the number of sampled patches. We believe that this occurs because the policy network samples more patches from the challenging and ambiguous cases to ensure that the classifier successfully predicts the label. On the other hand, it successfully learns when to sample \emph{no patches}. However, it samples no patch ($S$=0) $7\%$ of the time on average in comparison to sampling 4$\leq$S$\leq$7 $50\%$ of the time. Increasing the ratio for $S$=0 is a future work of this study. Finally, Fig.~\ref{fig:analysis_pn} (right) displays the probability of sampling a patch given its position. We see that the policy network learns to sample the central patches more than the peripheral patches as expected.

\section{Improving Run-time Complexity of BagNets}

\begin{figure}[!h]
\centering
\includegraphics[width=0.48\textwidth]{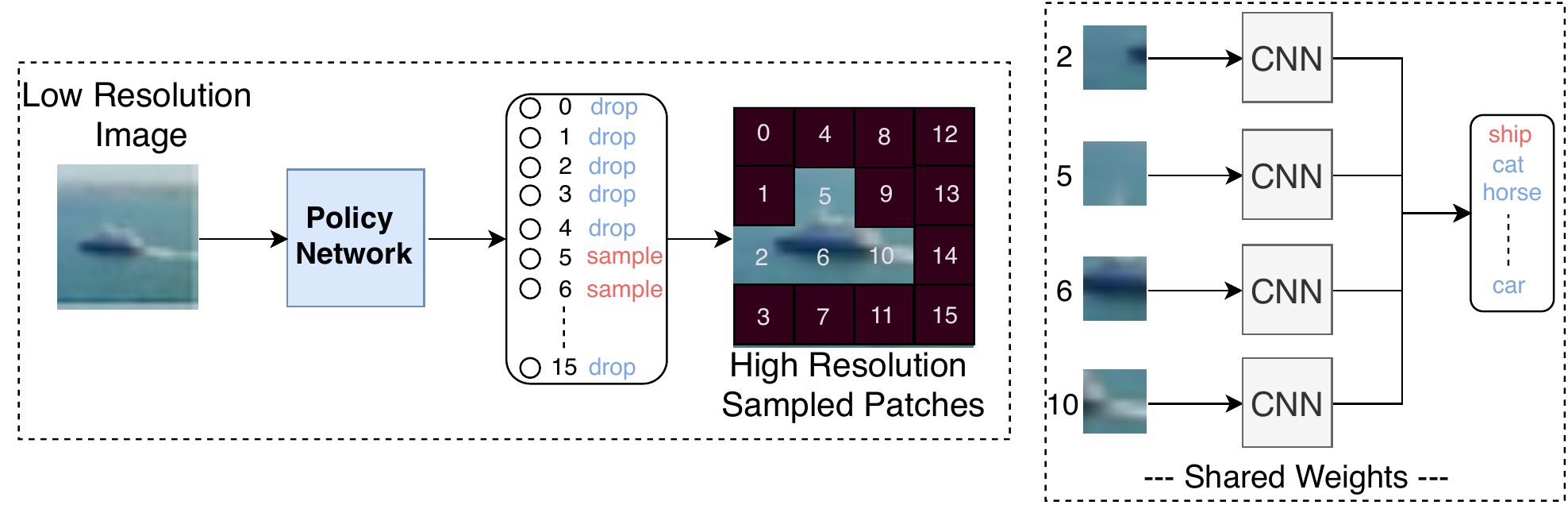}
\caption{Dynamic BagNet. The policy network processes LR image and sample HR patches to be processed independently by CNN. More details on BagNet can be found in~\cite{brendel2019approximating}.} 
\label{fig:bagnet_workflow}
\end{figure}

\begin{table}[!b]
\centering
\resizebox{0.91\columnwidth}{!}{%
\begin{tabular}{@{}llllll@{}}
\toprule
 & \multicolumn{1}{c}{\begin{tabular}[c]{@{}c@{}}Acc. (\%)\\ (Pt)\end{tabular}} & S & \multicolumn{1}{c}{\begin{tabular}[c]{@{}c@{}}Acc. (\%)\\ (Ft-1)\end{tabular}} & S & \multicolumn{1}{c}{\begin{tabular}[c]{@{}c@{}}Run-time. (\%)\\ (ms)\end{tabular}}\\ \midrule
BagNet (No Patch Drop)~\cite{brendel2019approximating} & 85.6 & 16 & 85.6 & 16 & 192\\
CNN (No Patch Drop) & 92.3 & 16 & 92.3 & 16  & 77\\\bottomrule
Fixed-H & 67.7 & 10 & 86.3 & 9 & 98\\
Fixed-V & 68.3 & 10 & 86.2 & 9 & 98 \\
Stochastic & 49.1 & 10 & 83.1 &  9 & 98\\
STN~\cite{ledig2017photo} & 67.5 & 10 & 86.8 & 9 & 112\\
\textbf{BagNet (PatchDrop)} & \textbf{77.4} & \textbf{9.5} & \textbf{92.7} & \textbf{8.5} & 98\\ \bottomrule
\end{tabular}}
\vspace{0.3em}
\caption{The performance of the PatchDrop and other models on improving BagNet on CIFAR10 dataset. We use a similar set up to our previous CIFAR10 experiments.}
\label{table:BagNet}
\end{table}

Previously, we tested PatchDrop on fMoW satellite image recognition task to reduce the financial cost of analyzing satellite images by reducing dependency on HR images while preserving accuracy. Next, we propose the use of PatchDrop to decrease the run-time complexity of local CNNs, such as BagNets. They have recently been proposed as a novel image recognition architecture~\cite{brendel2019approximating}. They run a CNN on image patches independently and sum up class-specific spatial probabilities. Surprisingly, the BagNets perform similarly to CNNs that process the full image in one shot. This concept fits perfectly to PatchDrop as it learns to select semantically useful local patches which can be fed to a BagNet. This way, the BagNet is not trained on all the patches from the image but only on \emph{useful patches}. By dropping redundant patches, we can then speed it up and improve its accuracy. In this case, we first train the BagNet on all the patches and pre-train the policy network on LR images (4$\times$) to learn patches important for BagNet. Using LR images and a shallow network (ResNet8), we reduce the run-time overhead introduced by the agent to $3\%$ of the CNN (ResNet32) using HR images. Finally, we jointly finetune (Ft-1) the policy network and BagNet. We illustrate the proposed conditional BagNet in Fig.~\ref{fig:bagnet_workflow}.

We perform experiments on CIFAR10 and show the results in Table~\ref{table:BagNet}. The proposed Conditional BagNet using PatchDrop improves the accuracy of BagNet by $7\%$ closing the gap between global CNNs and local CNNs. Additionally, it decreases the run-time complexity by $50\%$, significantly reducing the gap between local CNNs and global CNNs in terms of run-time complexity\footnote{The run-times are measured on Intel i7-7700K CPU@4.20GHz}. The increase in the speed can be further improved by running different GPUs on the selected patches in parallel at test time.

Finally, utilizing learned masks to avoid convolutional operations in the layers of global CNN is another promising direction of our work.~\cite{ghiasi2018dropblock} drops spatial blocks of the feature maps of CNNs in training time to perform stronger regularization than DropOut~\cite{srivastava2014dropout}. Our method, on the other hand, can drop blocks of the feature maps dynamically in both training and test time.

\begin{table}[!h]
\centering
\resizebox{0.9\columnwidth}{!}{\begin{tabular}{@{}llllll@{}}
\toprule
 & \begin{tabular}[c]{@{}c@{}}CIFAR10 (\%)\\ (ResNet32)\end{tabular} & \begin{tabular}[c]{@{}c@{}}CIFAR100 (\%)\\ (ResNet32)\end{tabular} & \begin{tabular}[c]{@{}c@{}}ImageNet (\%)\\ (ResNet50)\end{tabular} & \begin{tabular}[c]{@{}c@{}}fMoW (\%)\\ (ResNet34)\end{tabular} \\ \midrule
No Augment. & 92.3 & 69.3 & 76.5 & 67.3 \\
CutOut~\cite{devries2017improved} & 93.5 & 70.4 & 76.5 & 67.6 \\ 
PatchDrop & \textbf{93.9} & \textbf{71.0} & \textbf{78.1} & \textbf{69.6} \\
\bottomrule
\end{tabular}}
\caption{Results with different augmentation methods.}
\label{tab:hard_positive}
\end{table}
\section{Conditional Hard Positive Sampling}
 
 PatchDrop can also be used to generate hard positives for data augmentation. In this direction, we utilize the masked images, $\mathcal{X}_h^m$, learned by the policy network (Ft-1) to generate hard positive examples to better train classifiers. To generate conditional hard positive examples, we choose the number of patches to be masked, $M$, from a uniform distribution with minimum and maximum values of 1 and 4. Next, given $s_{p}$ by the policy network, we choose $M$ patches with the highest probabilities and mask them and use the masked images to train the classifier. Finally, we compare our approach to CutOut~\cite{devries2017improved} which randomly cuts/masks image patches for data augmentation. As shown in Table~\ref{tab:hard_positive}, our approach leads to higher accuracy in all the datasets when using original images, $\mathcal{X}_h$, in test time. This shows that the policy network learns to select informative patches.


%% file: conclusion.tex
\section{Conclusion}
In this study, we proposed a novel reinforcement learning setting to train a policy network to learn \emph{when} and \emph{where} to sample high resolution patches conditionally on the low resolution images. Our method can be highly beneficial in domains such as remote sensing where high quality data is significantly more expensive than the low resolution counterpart. In our experiments, on average, we drop a 40-60$\%$ portion of each high resolution image while preserving similar accuracy to networks which use full high resolution images in ImageNet and fMoW. Also, our method significantly improves the run-time efficiency and accuracy of BagNet, a patch-based CNNs. Finally, we used the learned policies to generate hard positives to boost classifiers' accuracy on CIFAR, ImageNet and fMoW datasets.